\documentclass[journal]{IEEEtran}

\ifCLASSINFOpdf
\else
\usepackage[dvips]{graphicx}
\fi
\usepackage{fixltx2e}
\usepackage{url}
\usepackage{multirow}
\usepackage{amsfonts}	
\usepackage{graphicx}
\usepackage{tikz}
\def\checkmark{\tikz\fill[scale=0.4](0,.35) -- (.25,0) -- (1,.7) -- (.25,.15) -- cycle;} 

\begin{document}
\title{Multiscaled Multi-Head Attention-based Video Transformer Network for Hand Gesture Recognition}

\author{Mallika Garg, Debashis Ghosh, \IEEEmembership{Senior Member, IEEE} and Pyari Mohan Pradhan, \IEEEmembership{Member, IEEE}
    \thanks{The authors are with the Department of Electronics and Communication Engineering, Indian Institute of Technology, Roorkee,  India-247667, (e-mail: mallika@ec.iitr.ac.in, debashis.ghosh@ece.iitr.ac.in, pyarimohan.pradhan@gmail.com).}}

\markboth{}
{Shell \MakeLowercase{\textit{et al.}}: Bare Demo of IEEEtran.cls for IEEE Journals}

\maketitle

\begin{abstract}
Dynamic gesture recognition is one of the challenging research areas due to  variations in pose, size, and shape of the signer's hand. In this letter, Multiscaled Multi-Head Attention  Video Transformer Network (MsMHA-VTN) for dynamic hand gesture  recognition is proposed. 	A  pyramidal hierarchy of  multiscale features is extracted using the transformer multiscaled head attention model. The proposed model employs different attention dimensions for each head of the transformer which enables it to provide attention at the multiscale level. Further,  in addition to single modality, recognition performance using multiple modalities is examined.  Extensive experiments  demonstrate the superior performance of the proposed MsMHA-VTN with an overall accuracy of  88.22\% and 99.10\% on NVGesture and Briareo datasets, respectively.  	
\end{abstract}	
\IEEEpeerreviewmaketitle
\section{Introduction}      
With advancements in deep learning, its applications in the fields of pattern recognition and human-computer interaction have gained much attention. This enables the automatic extraction of self-learned features thereby	reducing the need for manual extraction and selection of suitable discriminating features.  This advancement is more prominent in recognizing dynamic gestures  where spatio-temporal features are extracted. Earlier, Long Short-Term Memory (LSTM)-based techniques~\cite{27} were  used  for sequential data. Since then, several variants of LSTMs and Recurrent Neural Networks (RNN) have been developed. The effects of the attention mechanism in LSTM are further  explored in~\cite{28}. An attention-based Gated recurrent unit~\cite{29} and recurrent three-dimensional  convolutional neural network (R3DCNN)~\cite{10} have also been developed for extracting temporal features for dynamic gesture recognition. 

The attention mechanism decides and identifies the most important part of the gesture video sequence at each learning step. One such model which is based on the attention mechanism is the Transformer~\cite{30}. Transformers  basically use an encoder and decoder to sequentially model the input to an output sequence, which can be used as an alternative to RNN-based and LSTM-based models.	The encoder maps the input sequence and its positional encoding to some high-dimensional attention vector. Positional encoding may either be learned or fixed positional encoding. Few encoders are stacked together and the output of the final encoder is then fed to the decoder. The decoder finally maps this high-dimensional attention vector to the final decoding vector.

Initially, transformers were proposed for Sequence-to-Sequence translation in natural  language processing. Later, they were also applied to a sequence of image patches for image classification~\cite{31}.  Previous studies show that Vision Transformer (ViT) and Data efficient Image Transformer (Deit)~\cite{32} are two suitable models for using transformers in computer vision. A Transformer network can also be used for video recognition~\cite{33, 24}. In Video Transformer Networks (VTN), the spatial features from each frame are extracted using a pre-trained backbone network. Further, a transformer model that extracts multiscaled features using pooling in Multiscale Vision Transformers (MViT)  was proposed in~\cite{1}. An improved version of MViT~\cite{3} incorporates decomposed relative positional embedding and residual pooling connections which outperforms the MViT in terms of accuracy and computational overhead.

The transformer-based model has also been proposed for Dynamic Hand Gesture Recognition~\cite{7}. Frame-level features are extracted from the backbone network which is subsequently combined with the temporal information of the frame. The basic transformer model lacks ordering information as the video data is received in parallel. This is dealt with a Gated Recurrent Unit-Relative Sign Transformer (GRU-RST)~\cite{34} which uses relative positional embedding. With recent advancements, better recognition performances in the case of continuous gestures have been reported in~\cite{4} and~\cite{35}.

In this letter, a framework for the classification of dynamic hand gestures based on transformer architecture has been proposed. Single and multimodal inputs are used in the proposed method. Unlike the original transformer module, the proposed model employs multiscaled attention from different heads of the transformer in a pyramidal structure. 	

 Summarizing, the major contributions of this letter are:	
\begin{enumerate}
    \item A multiscaled multi-head attention to capture multiscale features in a transformer model  is proposed  in which a pyramid of scaled attention is designed. This significantly enhances the performance over the original basic transformer network~\cite{30} for dynamic gesture recognition.
    
    \item Single and multiple inputs from active data sensors are utilized for single-modal as well as multi-modal	dynamic gesture recognition. These inputs are RGB colors, depth,	infrared images, surface normals estimated from depth maps, and optical flow estimated from the RGB images.
    
    \item Efficacy of the proposed framework is validated on two publicly released datasets, viz., NVidia Dynamic Hand Gesture and Briareo data and the proposed model leads to state-of-the-art results, compared to existing methods.
\end{enumerate}

\section{MULTISCALED MULTI-HEAD  TRANSFORMER}
Let $X = [X_1,X_2,..,X_n]$ denotes the input set of $n$ gesture video samples, where $X_i$ represents the $i^{th}$ sample given in the form $X_i = \{x_{i1}, x_{i2},..,x_{it}\}$, where $x_{it}$ denotes the $t^{th}$ frame of the $i^{th}$ gesture video. The goal is to predict the class label  to which an input video sample may belong.

\subsection{Revisiting Multi-Head Attention}\label{formats}
The transformer model  presented in~\cite{30} uses scaled dot-product attention and multi-head attention along with addition of positional encoding to the input embedding. The attention comprises of three vectors, viz., Query ($\textbf{Q}$), Key ($\textbf{K}$) and Value ($\textbf{V}$), which are mapped to the final output. Accordingly, the output is computed as
\begin{equation}
\label{eqn:1}
Attention(\textbf{Q},\textbf{K},\textbf{V}) = softmax \left(\frac{\textbf{Q}\textbf{K}^T}{\sqrt{d_k}} \right)\textbf{V},
\end{equation}
where $d_k$ is the dimension of the key $\textbf{K}$ , and the query $\textbf{Q}$.

In the transformer model, instead of using single attention for all the three vectors $\textbf{Q}$, $\textbf{K}$ and $\textbf{V}$, having dimensions $d_{model}$,  the attention maps are linearly projected  onto $h$ different heads.  Conventionally, we take $d_{model}=512$ and $h=8$ for parallel attentions and the dimensions  for query, key, and value are taken as $d_k$, $d_k$ and $d_v = d_{model}/h = 64$, respectively. Subsequently, the scaled dot-product in (\ref{eqn:1}) is computed for each of the eight heads. The main aim of creating multi-head is that it represents the attention in different sub-spaces of same dimension which are learned in parallel. 

Now, let $W^\textbf{Q}_i \in \mathbb{R}^{d_{model} \times d_k}$ , $W^\textbf{K}_i \in \mathbb{R}^{d_{model} \times d_k}$, $W^\textbf{V}_i \in \mathbb{R}^{d_{model} \times d_v}$ are the respective learned weights for the three attention vectors \textbf{Q}, \textbf{K}, and \textbf{V}, corresponding to the $i^{th}$ head, and $W_O \in \mathbb{R}^{hd_v \times d_{model}}$ is the weight for the output multi-head attention. Then, the $i^{th}$ head may be defined as
\begin{equation}
\label{eqn:3}
head_i = Attention(\textbf{Q}W^\textbf{Q}_i,\textbf{K}W^\textbf{K}_i,\textbf{V}W^\textbf{V}_i).
\end{equation}	
Attention from different subspaces (heads)  in  (\ref{eqn:3})  are subsequently concatenated to get the final multi-head  attention output
\begin{equation}
\label{eqn:2}
MultiHead(\textbf{Q},\textbf{K},\textbf{V}) = Concat(head_1,..,head_h)W^O.
\end{equation}

\subsection{Multiscaled Multi-Head Attention}\label{4}
\begin{figure}				
\centerline{\includegraphics[width=8.3cm,height=4.2cm]{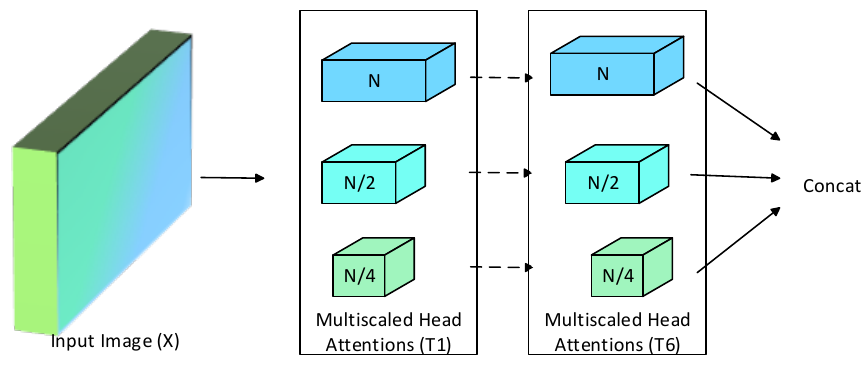}}
\caption{The proposed Multiscaled Multi-Head Attention Video Transformer Network (MsMHA-VTN) with 6 transformer stages (T1-T6) and 8 heads, where $N=L\times D$ is the dimension of the first head in each attention vector, $D$  and $L$  are the dimensions of the input tensor.  For convenience, pyramid scaling at each head by a factor of $1/2$ is shown only for three heads.}
\label{fig1}
\end{figure}

\begin{figure}				
    \centerline{\includegraphics[width=8cm,height=4.56cm]{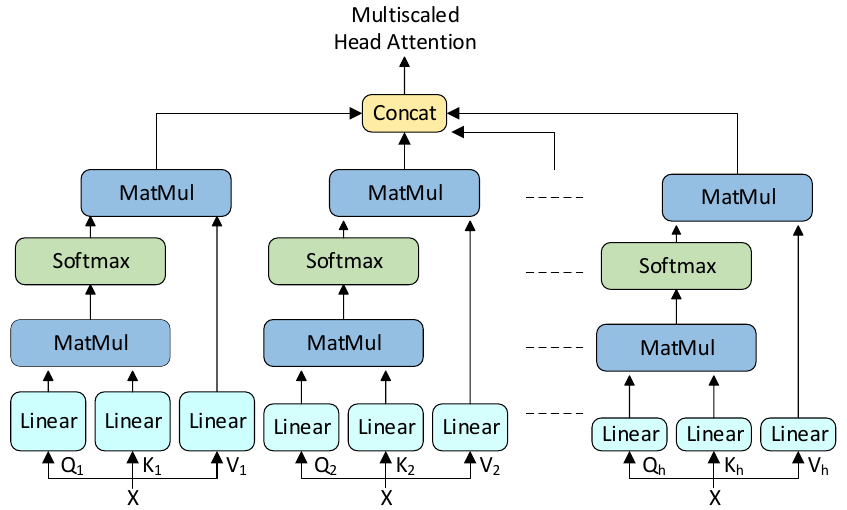}}
    \caption{The proposed Mutiscaled Multi-Head pyramid attention. Pyramid attention is an  attention mechanism that defines the length of the query, key and value in a pyramid pattern.  This allows to exploit the multi-scale information at different heads. Here, the size of linear block is kept varying w.r.t heads to show that $d_k$  decreases gradually.}
    \label{a}
 \end{figure}

In the proposed Multiscaled  Multi-Head Attention (MsMHA),  every head processes different dimensional subspace in contrast to~\cite{30} which considers same dimensional representations.  Thus, it extracts contextual information from different dimensional subspaces while providing  a flexible resolution of the given input data. Consequently, the proposed model learns a pyramid of attention with decreasing resolution arranged in a pyramidal pattern. Here, the vector dimensions of multi-head attention ($d_k$) decrease gradually for each head. This is achieved by scaling down each attention head tensor  by a factor of 2 as we move up the pyramid.  This allows a reduced resolution of receptive fields to capture features at different scales in each stage. 
The input to every  subspace is an image  $X \in \mathbb{R}^{L\times D}$, as shown in Fig.~\ref{fig1}, with a dimension of input tensor as $D$ and  $L$. In the Multi-Head Attention (MHA)~\cite{30}, the dimension, $d_k$  is same for all the heads while in the proposed MsMHA, the dimension of key, and query vectors  i.e., $d_k$ varies for each head at every stage, as shown in  Fig.~\ref{a}. 
Mathematically,  the input $X$ generates  query, key, and value attentions,  with linear operations such that $\textbf{Q} \in \mathbb{R}^{L\times D}$, $\textbf{K} \in \mathbb{R}^{L\times D}$ and $\textbf{V}  \in \mathbb{R}^{L\times D}$, respectively. The three attention vectors for the first head  are defined as
\begin{equation}
\label{eqn:4}
\textbf{Q}_{1D} = XW^\textbf{Q}_{1D}, ~\textbf{K}_{1D} = XW^\textbf{K}_{1D},  ~\textbf{V}_{1D} = XW^\textbf{V}_{1D},
\end{equation}
where the dimension of each attention vector  for the first head is $N=L\times D$, since the input dimension is $N$. For the remaining heads, the attention vectors are defined as, 
\begin{equation}
\label{eqn:5}
\textbf{Q}_{jM/2} = XW^\textbf{Q}_{jM/2}, ~\textbf{K}_{jM/2} = XW^\textbf{K}_{jM/2}, ~ \textbf{V}_{jM/2} = XW^\textbf{V}_{jM/2},
\end{equation} 
where $j \in \{2,h\}$ and $M$ is the dimension of the previous head tensor.  It may be  noted that the dimensions of query and key progressively  decreases with each head in the proposed transformer model,  as shown in Table~\ref{tab6}.		
\begin{table}
    \caption{Details of Multiscaled Head Attention Transformers (MsMHA-VTN) model for $h=8$.}
    \centering
    \begin{tabular}{ccc}
        \hline
        Head& Attention Tensor size wrt scale  \\ 
        \hline
        $head_1$& $L\times D$\\
        $head_2$& $L\times D/2$\\
        $head_3$& $L\times D/4$\\
        $head_4$& $L\times D/8$\\
        $head_5$& $L\times D/16$\\
        $head_6$& $L\times D/32$\\
        $head_7$& $L\times D/64$\\
        $head_8$& $L\times D/128$\\	\hline	
    \end{tabular}
    \label{tab6}
\end{table}
\subsection{Multiscaled Multi-Head Attention-based Transformer}\label{5}
The proposed MsMHA learns multiscale attentions from different heads and concatenates  them to get complete attention for a gesture.  This provides more focus on important features of a gesture with different size that varies from sample to sample due to varying pose, scale and size of the signers' hands.  Our proposed multiscaled multi-head attention repeats for six stages  which takes frame-level features as input from ResNet-18 model,  as developed in~\cite{7}. The ResNet-18 model is initialized with weights   pre-trained on the ImageNet database~\cite{111} in the same   way as  in~\cite{7}.
    
\subsection{Decision-Level Multi-Modal Fusion}\label{9}
The proposed transformer unimodal network can also be extended for multimodal gesture recognition by decision combining  using late fusion. Each of the different information cues available in the input data (color, depth, infrared, normals and optical flow)  from RGB-D sensors, is used to calculate classification score in a unimodal way. In late fusion,  the  decision-level multimodal fusion is  described as
\begin{equation}
\label{eqn:6}	
y = \arg\max_j \sum_{i}^m P(\omega_j|x_i),
\end{equation}
where $m$ is the total input streams, and  $P(\omega_j|x_i)$ is the classification score of the  $i^{th}$ input stream of a given input, which belongs to class $\omega_j$.
\section{Experimental Results}\label{6}
\subsection{Implementation Details}\label{7}
The proposed MsMHA-VTN model was implemented using PyTorch=1.6.0  with  Nvidia Tesla K80 GPU having 24GB RAM, CUDA 10.1  with cuDNN 8.1.1.  Initially, the  learning rate was set to $1e^{-4}$ which  decays at the $50^{th}$ and $75^{th}$ epoch, same as those used in~\cite{7}. Also, Adam optimizer was  used for training. 40 frames per gesture sequence  were given as input to the proposed model.

\begin{table}
\caption{results for different modalities on NVGesture dataset~\cite{10} . \# is the number of input modalities used.}
\centering
\begin{tabular}{ccccccc}
    \hline
    \multirow{2}{*}{\#}&\multicolumn{5}{c}{Input data}& \multirow{2}{*}{Accuracy} \\ \cline{2-6}
    &Color&Depth&IR&Normals&Optical flow&\\
    \hline
    \multirow{4}{*}{1} &\checkmark&&&&&81.42\%  \\
    && \checkmark&&&& 85.00\%\\
    && &\checkmark&&& 70.04\%\\
    && &&\checkmark&&\textbf{86.21}\%\\ \hline 
            
    \multirow{9}{*}{2} &\checkmark&\checkmark&&& &84.37\% \\
    &\checkmark&& \checkmark&&& 70.14\%\\
    &&\checkmark &\checkmark&& &83.00\%\\
    & \checkmark&&&\checkmark&& 86.25\%\\
    &&\checkmark &&\checkmark& &\textbf{87.90}\%\\
    && &\checkmark&\checkmark&&83.60\%\\ 
    &\checkmark&& &&\checkmark&85.30\% \\
    & &\checkmark&&&\checkmark&85.75\% \\
    &&&\checkmark&&\checkmark & 74.43\%\\ \hline
    
    \multirow{4}{*}{3} &\checkmark&\checkmark&\checkmark&&& 85.07\% \\
    &\checkmark& \checkmark&&\checkmark&& \textbf{87.82}\%\\
    &\checkmark& &\checkmark&\checkmark&& 85.10\%\\
    && \checkmark&\checkmark&\checkmark&&86.21\%\\ 
    \hline
    
    \multirow{5}{*}{4}&\checkmark&\checkmark &\checkmark&\checkmark&&87.81\%\\
    &\checkmark&\checkmark &\checkmark&&\checkmark&77.68\%\\ 
    &\checkmark&\checkmark &&\checkmark&\checkmark&\textbf{88.17}\%\\ 
    &\checkmark&&\checkmark&\checkmark&\checkmark&84.21\%\\ 
    &&\checkmark&\checkmark&\checkmark&\checkmark&85.15\%\\
    
    \hline
    5&\checkmark&\checkmark &\checkmark&\checkmark&\checkmark&\textbf{88.22}\%\\
    \hline
    
\end{tabular}
    \label{tab1}
\end{table}

\begin{table}
    \caption{comparison results for single modality on NVGesture dataset~\cite{10} }
    \centering
    \begin{tabular}{ccc}
        \hline
        Input modality&Method& Accuracy \\ 
        \hline
        \multirow{14}{*}{Color} &Spat. st. CNN~\cite{8} &54.60\%  \\
        &iDT-HOG~\cite{13}& 59.10\%\\
        &Res3ATN~\cite{12}& 62.70\%\\
        &C3D~\cite{15}&69.30\%\\ 
        &R3D-CNN~\cite{10}& 74.10\%\\
        &GPM~\cite{22}&75.90\% \\
        &PreRNN~\cite{21}&76.50\% \\
        &Transformer~\cite{7}& 76.50\%\\
        &I3D~\cite{17}&78.40\%\\
        &ResNeXt-101~\cite{14}& 78.63\%\\
        &MTUT~\cite{19}&81.33\%\\
    
        &NAS1~\cite{4}&83.61\% \\			
        &Human~\cite{10}&88.40\% \\
        &\textbf{MsMHA-VTN}& \textbf{81.42\%}\\
        \hline
        
        \multirow{11}{*}{Depth}& SNV~\cite{16}& 70.70\%\\
        &C3D~\cite{15}&78.80\%\\ 
        &R3D-CNN~\cite{10}& 80.30\%\\
        &I3D~\cite{17}&82.30\%\\
        &Transformer~\cite{7}&83.00\%\\
        &ResNeXt-101~\cite{14}& 83.82\%\\
        &PreRNN~\cite{21}&84.40\% \\
        &MTUT~\cite{19}&84.85\%\\
    
        &GPM~\cite{22}&85.50\% \\
        &NAS1~\cite{4}&86.10\% \\	
            &\textbf{MsMHA-VTN}& \textbf{85.00\%}\\
        
        \hline	
        
        \multirow{7}{*}{Optical flow}&iDT-HOF~\cite{13}& 61.80\% \\
        &Temp. st. CNN~\cite{8} & 68.00\%\\
        &Transformer~\cite{7}& 72.00\%\\
        &iDT-MBH~\cite{13} & 76.80\%\\
        &R3D-CNN~\cite{10} & 77.80\%\\
        &I3D~\cite{17} & 83.40\%\\
        &\textbf{MsMHA-VTN}&\textbf{85.30\%}\\
        \hline
        
        \multirow{2}{*}{Normals}&Transformer~\cite{7} &82.40\% \\
        &\textbf{MsMHA-VTN}& \textbf{86.21\%}\\ \hline
        
        \multirow{3}{*}{Infrared}&R3D-CNN~\cite{10}& 63.50\%\\
        & Transformer~\cite{7}&  64.70\% \\
        &\textbf{MsMHA-VTN}&\textbf{70.04\%}\\ \hline
        
    \end{tabular}
    \label{tab2}
\end{table}	
\begin{table}
    \caption{comparison results for multi-modalities on NVGesture dataset~\cite{10}. }
    \centering
    \begin{tabular}{ccc}
        \hline
        Method& Inputs modality& Accuracy  \\ 
        \hline
        Two-st. CNNs~\cite{8} & color + flow & 65.60\%\\
        \hline
        iDT ~\cite{13}& color + flow & 73.00\% \\
        \hline
        R3D-CNN~\cite{10} & color + flow & 79.30\%\\
        R3D-CNN~\cite{10} & color + depth + flow & 81.50\%\\
        R3D-CNN~\cite{10} & color + depth + ir & 82.00\%\\
        R3D-CNN~\cite{10} & depth + flow & 82.40\%\\
        R3D-CNN~\cite{10} & all & 83.80\%\\
        \hline
        MSD-2DCNN~\cite{22}&color+depth&84.00\% \\
        
        \hline
        8-MFFs-3f1c\cite{18}&color + flow& 84.70\%\\
        \hline
        STSNN~\cite{9}&color+flow& 85.13\%\\
        \hline
        PreRNN~\cite{21}& color + depth&85.00\% \\
        \hline
        
        I3D~\cite{17}& color + depth &83.80\%\\
        I3D~\cite{17}& color + flow &84.40\%\\
        I3D~\cite{17}& color + depth + flow &85.70\%\\
        
        \hline
        GPM~\cite{22}& color + depth&86.10\% \\
        \hline
        MTUT\textsubscript{RGB-D}~\cite{19}& color + depth& 85.50\%\\
        MTUT\textsubscript{RGB-D+flow}~\cite{19}& color + depth& 86.10\%\\
        MTUT\textsubscript{RGB-D+flow}~\cite{19}& color + depth + flow& 86.90\%\\
        \hline
        
        Transformer~\cite{7}& depth + normals &87.30\%\\
        Transformer~\cite{7}& color + depth + normals+ir& 87.60\%\\
        \hline
        NAS2~\cite{4}& color + depth&86.93\% \\
        NAS1+NAS2~\cite{4} &color + depth&88.38\% \\
        \hline
        
        \textbf{MsMHA-VTN}&\textbf{color + normals}&\textbf{87.90\%}\\	
        \textbf{MsMHA-VTN}&\textbf{depth + normals + ir + flow}&\textbf{88.17\%}\\
        \multirow{2}{*}{\textbf{MsMHA-VTN}}&{\textbf{color + depth + normals +}}&\multirow{2}{*}{\textbf{88.22\%}}\\ 
        &\textbf{ir + flow}& \\
        \hline	
    \end{tabular}
    \label{tab3}
\end{table}

\begin{table}
    \caption{ results for different modalities on Briareo dataset~\cite{20}. \# is the number of input modalities used.}
    \centering
    \begin{tabular}{ccccccc}
        \hline
        \multirow{2}{*}{\#}&\multicolumn{5}{c}{Input data}& \multirow{2}{*}{Accuracy} \\ \cline{2-6}
        &Color&Depth&IR&Normals&Optical flow&\\
        \hline
        \multirow{4}{*}{1} &\checkmark&&&&&98.15\%  \\
        && \checkmark&&&& 94.44\%\\
        && &\checkmark&&& 96.29\%\\
        && &&\checkmark&&\textbf{98.67}\%\\ \hline

        \multirow{9}{*}{2} &\checkmark&\checkmark&&& &97.22\% \\
        &\checkmark&& \checkmark&&& 98.14\%\\
        &&\checkmark &\checkmark&& &96.30\%\\
        & \checkmark&&&\checkmark&& 98.15\%\\
        &&\checkmark &&\checkmark& &97.22\%\\
        && &\checkmark&\checkmark&&97.22\%\\ 
        &\checkmark&& &&\checkmark&\textbf{98.75}\% \\
        & &\checkmark&&&\checkmark&97.30\% \\
        &&&\checkmark&&\checkmark & 97.27\%\\ \hline
        
        \multirow{4}{*}{3} &\checkmark&\checkmark&\checkmark&&& 99.07\% \\
        &\checkmark& \checkmark&&\checkmark&& 98.15\%\\
        &\checkmark& &\checkmark&\checkmark&& \textbf{99.10}\%\\
        && \checkmark&\checkmark&\checkmark&&99.07\%\\ 
    
        \hline
        
        \multirow{5}{*}{4}&\checkmark&\checkmark &\checkmark&\checkmark&&\textbf{98.61}\%\\
        &\checkmark&\checkmark &\checkmark&&\checkmark&97.68\%\\ 
        &\checkmark&\checkmark &&\checkmark&\checkmark&97.22\%\\ 
        &\checkmark&&\checkmark&\checkmark&\checkmark&\textbf{98.61}\%\\ 
        &&\checkmark&\checkmark&\checkmark&\checkmark&98.14\%\\
                    
        \hline
        5&\checkmark&\checkmark &\checkmark&\checkmark&\checkmark&\textbf{98.61}\%\\
        \hline
        
\end{tabular}
    \label{tab4}
\end{table}	
\begin{table}
    \caption{comparison results for different modalities on Briareo dataset~\cite{20}.}
    \centering
    \begin{tabular}{ccc}
        \hline
        Method& Inputs modality& Accuracy \\ 
        \hline
        C3D-HG~\cite{20}& color& 72.20\%\\
        C3D-HG~\cite{20}& depth& 76.00\%\\
        C3D-HG~\cite{20}& ir& 87.50\%\\
        
        LSTM-HG~\cite{20}&3D joint features &94.40\%\\
        \hline
        NUI-CNN~\cite{11}& depth + ir& 92.00\%\\
        NUI-CNN~\cite{11}& color + depth + ir& 90.90\%\\
        
        \hline
        Transformer~\cite{7}& normals& 95.80\%\\
        Transformer~\cite{7}& depth + normals &96.20\%\\
        Transformer~\cite{7}&ir + normals &97.20\%\\
        \hline	
        \textbf{MsMHA-VTN} & \textbf{normals}&\textbf{98.67} \%\\
        \textbf{MsMHA-VTN} & \textbf{color + optical} &\textbf{98.75}\%\\
        \textbf{MsMHA-VTN} &\textbf{color + ir + normals} &\textbf{99.10}\%\\
        \hline
    \end{tabular}
    \label{tab5}
\end{table}

\subsection{Results and Discussion}
 \textbf{NVGesture:}
 Table~\ref{tab1} shows the performance analysis in case of NVGesture dataset for different  modalities  which is achieved using the late fusion discussed in Section~\ref{9}. Thus, we observe that the model achieves state-of-the-art results when depth or normal is given as input. This shows that normals driven from depth can be used as a good representation of hand gestures. Where two and three modalities were combined, the accuracy remained nearly same, around 87.90\% and 87.22\%, respectively. This further improved to 88.17\% when four streams  of input were used. The highest accuracy of 88.22\% was achieved when color, depth, normals, infrared (IR) and optical flow images were all given as inputs.  
 
 Table~\ref{tab2} and Table~\ref{tab3} present comparative performance results when compared to some other approaches with  NVGesture dataset. It is observed that the proposed model yields comparable or even better results than other methods when only color and normals  or all five inputs are utilized.

\textbf{Briareo:} 
Table~\ref{tab4} presents the results for single and multimodal approaches applied to Briareo dataset. Recognition accuracy as high  as  98.67\% was obtained for surface normals compared to other modalities. Therefore, it may be inferred that normals which are driven from depth	contain more information about the gesture pattern than other modalities. On the other hand, it is observed that the combination of two modalities (color and optical flow) increases the accuracy to 98.75\%. However, combination of more than three modalities did not provide any improvement. The highest accuracy of 99.10\% was achieved  when  color, infrared and surface normals were fused. 

Table~\ref{tab5} presents the comparative performance   results for different modalities on Briareo dataset. It is observed that the proposed model outperforms the state-of-the-art methods with 99.10\% accuracy. Also, it is observed that our model performed better than other methods even when only surface normals were used for recognition. This shows that data from only one device is sufficient to achieve comparable results with our proposed MsMHA-VTN model.

\section{Conclusion}
A multiscaled head attention for transformer model is proposed in this work. The proposed model captures the contextual information at different levels to provide more focus on multiscale features at each stage. Multiscale features at each head follows a pyramid hierarchy which replicates at every stage in the transformer model. Analyzing the results, as obtained in our experiments, it may be concluded that the proposed MsMHA-VTN is able to learn multiscale attentions that significantly improve the performance of dynamic hand gesture recognition compared to some of the existing methods, particularly the original transformer model.

\end{document}